\documentclass[10pt,twocolumn,letterpaper]{article}

\usepackage{cvpr}              

\usepackage[dvipsnames]{xcolor}

\usepackage{bm}

\definecolor{cvprblue}{rgb}{0.21,0.49,0.74}
\usepackage[pagebackref,breaklinks,colorlinks,citecolor=cvprblue]{hyperref}

\def\paperID{8703} 
\def\confName{CVPR}
\def\confYear{2024}

\title{HOISDF: Constraining 3D Hand-Object Pose Estimation with Global Signed Distance Fields}

\author{Haozhe Qi
~~~~~~~~
   Chen Zhao 
   ~~~~~~~~
   Mathieu Salzmann 
   ~~~~~~~~
   Alexander Mathis \\
   École Polytechnique Fédérale de Lausanne (EPFL),
   Switzerland \\
   \texttt{{[first name].[surname]}@epfl.ch}
}

\begin{document}
\maketitle
\begin{abstract}
Human hands are highly articulated and versatile at handling objects. Jointly estimating the 3D poses of a hand and the object it manipulates from a monocular camera is challenging due to frequent occlusions. Thus, existing methods often rely on intermediate 3D shape representations to increase performance. These representations are typically explicit, such as 3D point clouds or meshes, and thus provide information in the direct surroundings of the intermediate hand pose estimate. To address this, we introduce HOISDF, a Signed Distance Field (SDF) guided hand-object pose estimation network, which jointly exploits hand and object SDFs to provide a global, implicit representation over the complete reconstruction volume. Specifically, the role of the SDFs is threefold: equip the visual encoder with implicit shape information, help to encode hand-object interactions, and guide the hand and object pose regression via SDF-based sampling and by augmenting the feature representations. We show that HOISDF achieves state-of-the-art results on hand-object pose estimation benchmarks (DexYCB and HO3Dv2). Code is available at~\href{https://github.com/amathislab/HOISDF}{https://github.com/amathislab/HOISDF}.
\end{abstract}
\vspace{-.75cm}

\section{Introduction}
\label{sec:intro}

Pose estimation during hand-object interaction from a single monocular view can contribute to widespread applications, e.g., in augmented reality~\cite{chen2019overview}, robotics~\cite{billard2019trends,chiappa2024acquiring}, human-computer interaction~\cite{ren2020review}, and neuroscience~\cite{mathis2020deep}. Many excellent 3D hand~\cite{zimmermann2019freihand, lin2021end, xiong2019a2j, rong2021frankmocap} and object~\cite{chen2022epro, peng2019pvnet, he2020pvn3d} pose estimation algorithms have been developed. However, due to severe occlusion, they can easily fail during hand-object interactions. This has led to the emergence of dedicated hand-object interaction datasets~\cite{hampali2020honnotate, chao2021dexycb, liu2022hoi4d, hasson2019learning}, and subsequently joint hand-object pose estimation has drawn increasing attention. Despite much progress, most methods still struggle when the hand or object is heavily occluded~\cite{park2022handoccnet, hampali2022keypoint, li2021artiboost, chen2021joint,tse2022collaborative, wang2023interacting, hasson2021towards}. We argue that this limitation is rooted in the way 3D shape information is embedded in these algorithms.

\begin{figure*}[!tp]
\begin{center}
\includegraphics[width=.9\textwidth]{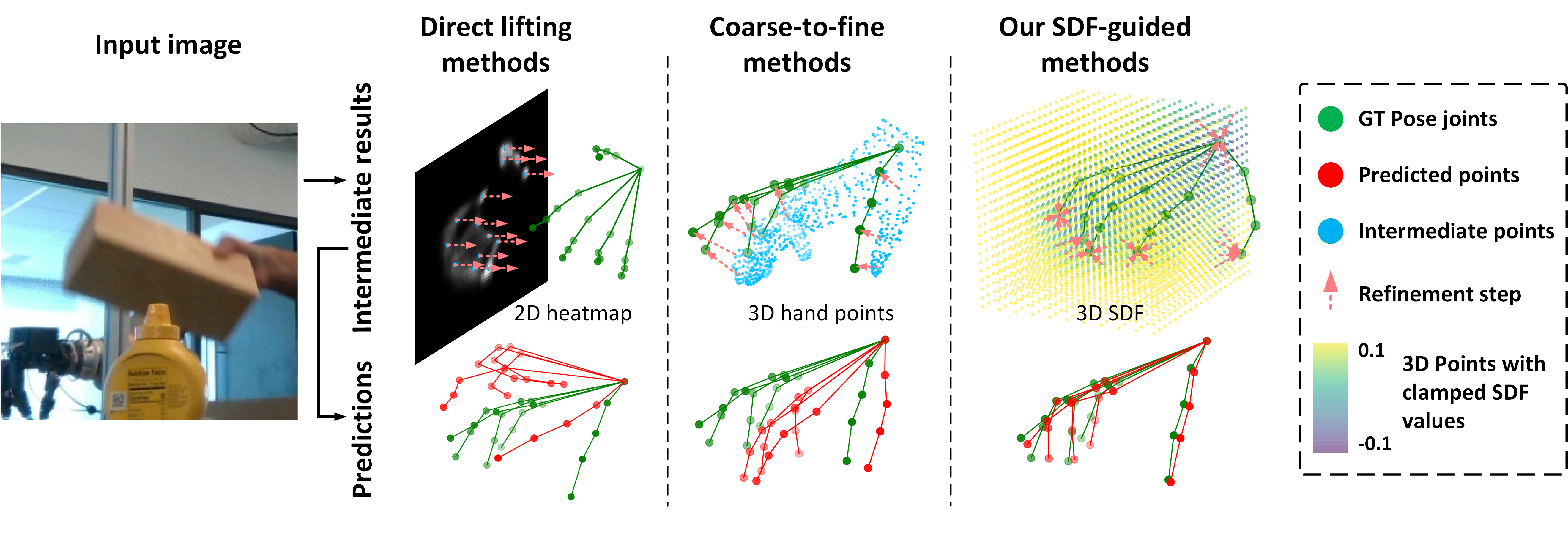}
\end{center} 
\vspace{-.5cm}
\caption{{\bf Conceptual advantage of the SDF-guided model over existing approaches.} Our model utilizes Signed Distance Fields (SDF) to provide global and dense constraints for hand-object pose estimation. In contrast to direct lifting and coarse-to-fine methods, which struggle to refine poor initial predictions, the distance field yields global cues not limited to areas near an initial prediction.}
\label{fig:fig1}
\vspace{-0.5cm}
\end{figure*}

In essence, existing methods can be classified into two approaches: Direct lifting and coarse-to-fine methods (see  Figure~\ref{fig:fig1}). Direct lifting methods first filter 2D image features according to the pixel positions of the hand and object and then use the remaining features to make predictions~\cite{park2022handoccnet, hampali2022keypoint, li2021artiboost, chen2021joint, lin2023harmonious}. These methods do not utilize explicit 3D intermediate representations and rely entirely on the network to learn the mapping from 2D image to 3D pose. Coarse-to-fine techniques make an initial prediction from the 2D image and improve upon it with a refinement network~\cite{tse2022collaborative, wang2023interacting, hasson2021towards, chen2022alignsdf, chen2023gsdf}. The intermediate representations can either be hand joints~\cite{chen2023gsdf, chen2022alignsdf} or hand vertices~\cite{tse2022collaborative, wang2023interacting, hasson2021towards}, which can be interpreted as explicit shape representations. Although these representations can incorporate 3D shape information, we argue that implicit shape representations in the form of signed distance fields (SDFs) offer more effective 3D shape information for subsequent computations.

To achieve this, we introduce HOISDF (a Hand-Object Interaction pose estimation network with Signed-Distance Fields), which uses SDFs to guide the 3D hand-object pose estimation in a global manner (Figure~\ref{fig:fig1}). HOISDF consists of two sequential components: a module learning to predict the signed distance field, and a module that performs pose regression that is field-guided (Figure~\ref{fig:architecture}). The signed distance field learning module regresses the hand and object signed distance fields based on the image features. The module is encouraged to focus on capturing global information (e.g., rough hand/object shape, global rotation and translation) by regressing signed distances in the original camera space, since we believe global plausibility is more important in the intermediate stage, while fine-grained details can be recovered in the later stages. To effectively leverage the dense field information, our field-guided pose regression module effectively uses the learned field information to (i) sample informative query points, (ii) augment the image features for those points, (iii) gather cross-target (i.e., hand-to-object or object-to-hand) cues to reduce the influence of mutual occlusion, and (iv) combine the point features together to estimate the hand and object poses. 

Overall, HOISDF can be trained in an end-to-end manner. We achieve state-of-the-art results on the DexYCB and HO3Dv2 datasets, corroborating the benefits of using SDFs as global constraints for hand-object pose estimation and the effectiveness of our approach to exploiting the field information. Altogether, our main contributions are:

\begin{itemize}
    \item We introduce a hand-object pose estimation network that uses signed distance fields (HOISDF) to introduce implicit 3D shape information
    \item We develop a new signed-distance field-guided pose regression module to effectively integrate the relevant parts of the global field information for hand and object pose estimation.
\end{itemize}

\section{Related Work}

\label{sec:rel_work}

\subsection{3D Hand-Object Pose Estimation}

Recently, joint hand-object pose estimation has drawn increasing research interest~\cite{lepetit2020recent}, and many hand and object interaction datasets have been developed~\cite{hampali2020honnotate, chao2021dexycb, liu2022hoi4d, hasson2019learning}. The current methods can be divided into direct lifting techniques and coarse-to-fine strategies. Among the former, Chen \emph{et al.} \cite{chen2021joint} fused hand and object features with sequential LSTM models. Hampali \emph{et al.} \cite{hampali2022keypoint} extracted 2D keypoints and sent them to a transformer architecture to find the correlation with the 3D poses. Li \emph{et al.} \cite{li2021artiboost} proposed a data synthesis pipeline that can leverage the training feedback to enhance hand object pose learning. Lin \emph{et al.} \cite{lin2023harmonious} proposed to learn harmonious features by avoiding hand-object competition in middle-layer feature learning. For the coarse-to-fine methods, Hasson \emph{et al.} \cite{hasson2021towards} obtained  initial hand and object meshes and optimized them with interaction constraints. Tse \emph{et al.} \cite{tse2022collaborative} used an attention-guided graph convolution to iteratively extract features from the previous hand-object estimates. Wang \emph{et al.} \cite{wang2023interacting} designed a dense mutual attention module to explore the relations from the initial hand-object predictions. We build on those methods but, in contrast, focus on implicit 3D shape information by learning SDFs, which provide global, dense constraints to guide the pose predictions. 

\subsection{Distance Fields in Hand-Object Interactions} 
Unlike explicit representations such as point clouds and meshes, neural distance fields provide a continuous and differentiable implicit representation that encodes the 3D shape information into the network parameters. Given a 3D query point, a neural distance field outputs the signed or unsigned distance from this point to the object surface. Neural distance fields have been widely used in 3D shape reconstruction and representation~\cite{park2019deepsdf, chibane2020neural, ma2020neural, yen2021inerf,or2022stylesdf}. Recently, SDFs have also been exploited in the context of hand-object interaction. In particular, Karunratanakul \emph{et al.} \cite{karunratanakul2020grasping} proposed to jointly model the hand, the object, and contact areas using an SDF. Ye \emph{et al.} \cite{ye2022s} used an SDF and the predicted hand to infer the shape of a hand-held object. Chen \emph{et al.} \cite{chen2022alignsdf} pre-aligned the 3D space with hand-object global poses to support the SDF prediction. Chen \emph{et al.} \cite{chen2023gsdf} further used entire kinematic chains of local pose transformations to obtain finer-grained alignment. However, those methods mainly use SDF as the endpoint of the model to directly reconstruct 3D meshes instead of using SDF as an intermediate representation. Here we explore how SDFs as an intermediate representations can guide subsequent pose estimation. Our experiments clearly demonstrate the benefits of our approach.

\begin{figure*}[!tp]
\begin{center}
\includegraphics[width=\textwidth]{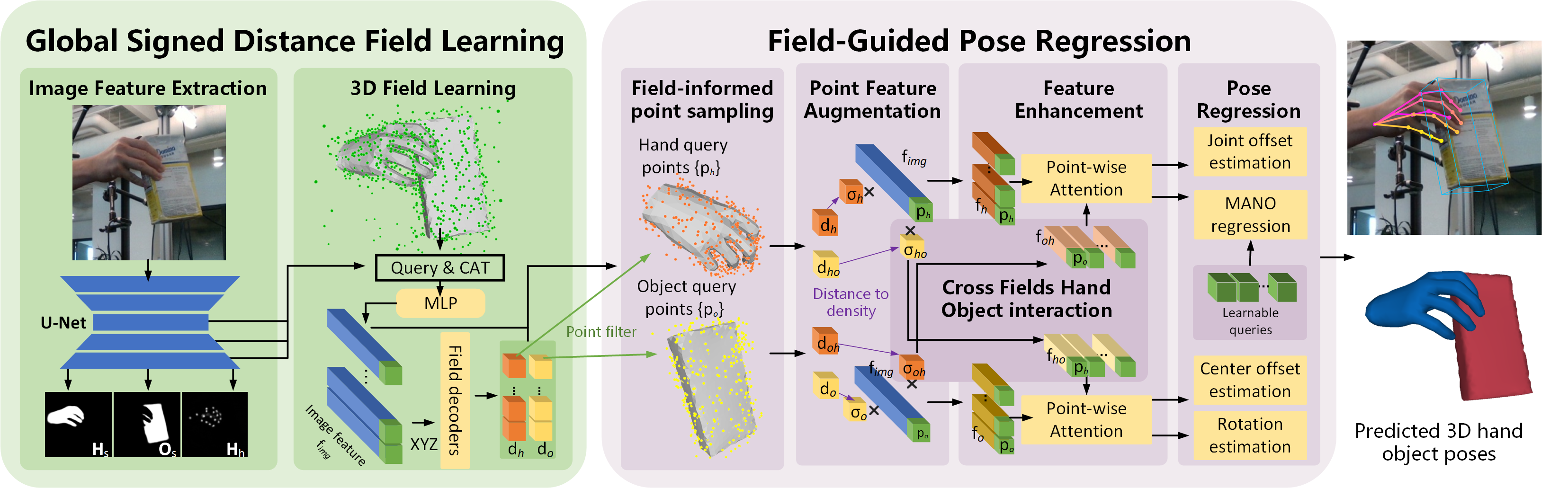}
\end{center} 
\vspace{-.5cm}
\caption{{\bf Overall pipeline of HOISDF.} HOISDF has two parts: A global signed distance field learning module and a field-guided pose regression module. The global signed distance field learning module regresses the hand object signed distances as the intermediate representation and encodes the 3D shape information into the image backbone through implicit field learning. The field-guided pose regression module uses global field information to filter and augment the point features as well as guiding hand-object interaction. Those enhanced point features are then sent to regress hand and object poses using point-wise attention.}
\label{fig:architecture}
\vspace{-.5cm}
\end{figure*}

\subsection{Attention-based Methods}

Attention mechanisms~\cite{vaswani2017attention} have been wildly successful in machine learning \cite{carion2020end, brown2020language, he2022masked, chiappa2022dmap,feichtenhofer2022masked} due to their effectiveness at exploiting long-range correlation. In the context of modeling hand-object relationships, Hampali et al. \cite{hampali2022keypoint} propose modeling correlations between 2D keypoints and 3D hand and object poses using cross attention. Tze \emph{et al.} \cite{tse2022collaborative} design an attention-guided graph convolution network to capture hand and object mesh information dynamically. Wang \emph{et al.} \cite{wang2023interacting} propose to exploit mutual attention between hand and object vertices to learn interaction dependencies. By contrast, our HOISDF applies attention across field-guided query points to mine the global 3D shape consistency context and cross-attend between hand and object.

\section{HOISDF}

We propose Hand-object Pose Estimation with Global Signed Distance Fields (HOISDF), a joint hand-object pose estimation model that leverages global shape constraints from a signed distance field. HOISDF comprises two components: A global signed distance field learning module and a field-guided pose regression module (Figure~\ref{fig:architecture}). Both components benefit from the robust 3D shape information modeled with the SDF and the whole architecture is trained end-to-end.

\begin{figure*}[!ht]
\begin{center}
\includegraphics[width=\textwidth]{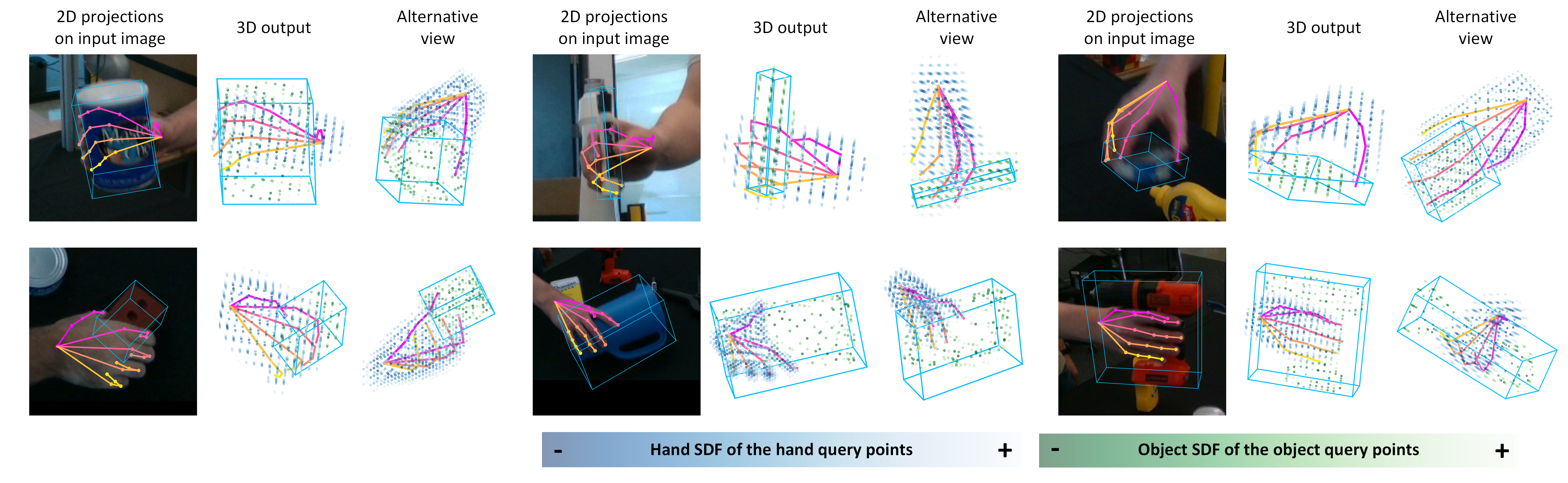}
\end{center} 
\vspace{-.5cm}
\caption{{\bf Visualization of the intermediate query points on DexYCB testset.} The darkness of the query points reflects the predicted distance from the query point to the hand (in blue) and object (in green) surfaces. The intermediate SDF representations can capture the GT 3D hand and object shapes. HOISDF effectively uses the robust global clues from SDFs to deal well with various objects and hand movements as well as their mutual occlusions.}
\label{fig:intermediate_dexycb}
\vspace{-0.5cm}
\end{figure*}

\subsection{Global Signed Distance Field Learning}
\label{sec:global_sdf}

We simultaneously learn hand and object signed distance fields (SDFs) with the following rationale: i) An SDF implicitly represents 3D shape with the model parameters; the implicit learning procedure can thus propagate 3D shape information to the feature extraction module. ii) Jointly learning hand and object fields allows the model to encode their mutual constraints. Meanwhile, since we predict hand and object signed distances in the initial stage as intermediate representations, we encourage our SDF learning module to focus more on global plausibility rather than local fine-grained details. Below, we describe the image feature extraction and the SDF learning in detail.

{\bf Image Feature Extraction.} For extracting hierarchical features $\textbf{F}$, we use a standard encoder-decoder architecture, specifically a U-Net~\cite{ronneberger2015u,he2016deep,hampali2022keypoint}. Following standard practice~\cite{hampali2022keypoint, wang2023interacting, lin2023harmonious}, we regress 2D predictions (a single channel heatmap~\cite{hampali2022keypoint} and hand/object segmentation masks with loss $\mathcal{L}_{img}$, see Supp. Mat.~\ref{sec:architecture-details} for details) to enable the model to represent hand-object interaction at the 2D image level. 

\label{sec:field}
{\bf 3D Signed Distance Field Learning.} With the extracted image features, the SDF module learns the continuous mapping from a 3D query point $\textbf{p} \in \mathbb{R}^3$ to the shortest signed distances between $\textbf{p}$ and the hand/object surfaces. Compared to \cite{chen2022alignsdf, chen2023gsdf}, we directly learn SDFs in the original space without rotating to canonical spaces using pose predictions. Our SDF module will consequently focus on the global information (e.g., general shape, location and global rotation) of the hand and object.

Specifically, given a 3D query point $\textbf{p} \in \mathbb{R}^3$, we project it to the 2D image space to compute the pixel-aligned image features \cite{hampali2022keypoint, wang2023interacting, chen2023gsdf, xie2022chore} extracted by the U-Net decoder $\{\textbf{F}_{dec}^{i}\}$, where $i \in \mathcal{X}$ indexes over the hierarchical decoder levels of the U-Net. We then concatenate the queried image features and pass them to a Multilayer Perceptron (MLP) to obtain a feature vector
\begin{align}
\label{pixel_feature}
\textbf{f}_{img}= \text{MLP}(\oplus_{i \in \mathcal{X}} \textbf{F}_{dec}^{i}(\pi_{3D \rightarrow 2D})),
\end{align}
where $\pi_{3D \rightarrow 2D}$ represents the projection and interpolation operation, $\oplus$ indicates the concatenation of all the hierarchical pixel-aligned image features, and $\mathcal{X}$ is the set of hierarchical features.

To emphasize the importance of $\textbf{p}$, we expand the coordinate representation by a Fourier Positional Encoding~\cite{mildenhall2021nerf} into a vector $\textbf{f}_{pos}$. We then concatenate the triplet $\textbf{p}$, $\textbf{f}_{pos}$ and $\textbf{f}_{img}$ together and pass them to the hand SDF decoder $\mathbb{SDF}_h$ and the object SDF decoder $\mathbb{SDF}_o$. This can be expressed as
\begin{align}
\label{equ:sdf_reg}
\textbf{f}_{sdf} & = \textbf{p}\oplus\textbf{f}_{pos}\oplus\textbf{f}_{img}, \\
d_{h} & =\mathbb{SDF}_h(\textbf{f}_{sdf}), \\
d_{o} & =\mathbb{SDF}_o(\textbf{f}_{sdf}).
\end{align}

Here, $d_{h}$ is the shortest distance from $\textbf{p}$ to the hand mesh surface, and $d_{o}$ is the shortest distance from $\textbf{p}$ to the object mesh surface; $d_{h}$ and $d_{o}$ will be positive if they are outside the surface and negative otherwise. The field decoders $\mathbb{SDF}_h$, and $\mathbb{SDF}_o$ are all 3-layer MLPs with \emph{tanh} activation in the last layer \cite{karunratanakul2020grasping}.

During training, we sample $N_s$ 3D query points, ensuring that most points are sampled near the hand and object mesh surfaces. We pre-compute the ground-truth distances from the query point to the hand and object surfaces and use the smooth-L1 loss~\cite{ren2015faster} to supervise the learning of $d_{h}$ and $d_{o}$. We sum the losses together and refer to the resulting loss as $\mathcal{L}_{sdf}$.

\subsection{Integrating Field Information: Field-guided Pose Regression}
\label{sec:pose_attention}

After the field learning module, we aim to use the learned fields to predict the hand and object poses. However, effectively using the field information is non-trivial: i) The field information is implicitly encoded in the model parameters; we can only read the field information at a specific location by sending a query point into the network; ii) The resulting signed distance at a certain query point is just a scalar distance, which on its own provides only a weak link with the pose prediction; iii) How to explicitly model the hand-object interaction using SDF is unclear. To address these challenges, we hence introduce the field-guided pose regression module described below.

\subsubsection{Field-informed Point Sampling} 
\label{sec:sampling}

To address the first problem, we propose a point-sampling strategy that aims to extract the most helpful field information while querying only a few points. It builds on the assumption that the query points near the ground-truth surface are the most informative ones. As such, during inference, we voxelize the 3D space with $N_v$ bins, which gives us $N_v^3$ query points. We first use the hand and object bounding boxes to filter the points in 2D space. Then, we send the remaining points into $\mathbb{SDF}_h$ and $\mathbb{SDF}_o$ and sort them according to the obtained hand and object signed distances separately. We sample $N_v^2/n_h$ hand query points and $N_v^2/n_o$ object query points with the lowest absolute hand distance and object distance, respectively. Here, $n_h$ and $n_o$ are two positive hyperparameters controlling the number of samples. Since we can access the ground-truth mesh during training, we directly sample $N_h$ hand query points near the hand mesh and $N_o$ object query points near the object mesh (with an absolute distance smaller than 4cm) for speed and memory optimization (2x faster). Towards the end of training, we also sample points with the same strategy as during testing to learn the point distribution. We will show the effectiveness of our proposed sampling strategy in Sec.~\ref{sec:ablation_representation}.

\subsubsection{Field-based Point Feature Augmentation}
\label{sec:fea_reg}

To address the second problem, given a sampled hand query point $\textbf{p}_h$, we convert $d_{h}$ to the volume density $\sigma_{h}=\alpha^{-1}sigmoid(-d_{h}/\alpha)$, where $\alpha$ is a learnable parameter to control the tightness of the density around the surface boundary. This is motivated by the strategy used in StyleSDF~\cite{or2022stylesdf} for image rendering, but here we use it for the purpose of feature augmentation. We then multiply $\sigma_{h}$ with $\textbf{f}_{img}$. The field information will thus influence the whole feature representation. $\textbf{p}_h$ and its positional encoding $\textbf{f}_{pos}$ discussed in Sec.~\ref{sec:field} are also concatenated to further augment the point feature. The final hand query point feature $\textbf{f}_{h}$ is obtained as
\begin{align}
\label{equ:feature_h}
\textbf{f}_{h} = \textbf{p}_h\oplus\textbf{f}_{pos}\oplus(\textbf{f}_{img}\cdot \sigma_{h}).
\end{align}

For a sampled object query point $\textbf{p}_o$, the object query point feature $\textbf{f}_{o}$ is obtained in an analogous way (i.e., augmenting the feature by the volume density $\sigma_{o}$ based on object SDF $d_{o}$).

\subsubsection{Cross Fields Hand-Object Interaction} 
\label{sec:cross_interaction}

Since we use the shared image backbone to learn the hand-object SDFs jointly, hand-object relations can be implicitly modeled during implicit field learning. Here, we aim to model the hand-object interaction explicitly to better deal with the mutual occlusions. Intuitively, the hand-object contact areas are highly informative about the object/hand pose. Therefore, we augment the hand/object query points with the object/hand SDFs, respectively, to serve as interaction cues (Fig.~\ref{fig:architecture}). Specifically, for a sampled object query point $\textbf{p}_o$, we send it to the hand SDF decoder $\mathbb{SDF}_h$ to obtain the cross-hand signed distance $d_{oh}$. $d_{oh}$ is then converted to the volume density $\sigma_{oh}$ and used to augment the queried image feature $\textbf{f}_{img}$ similarly to Sec.~\ref{sec:fea_reg}. The final cross-hand query point feature $\textbf{f}_{oh}$ is obtained as
\begin{align}
\label{equ:feature_oh}
\textbf{f}_{oh} = \textbf{p}_o\oplus\textbf{f}_{pos}\oplus(\textbf{f}_{img}\cdot \sigma_{oh}).
\end{align}
$\textbf{f}_{oh}$ will serve as object cues for hand pose estimation. A $\textbf{p}_o$ with smaller $d_{oh}$ will play a bigger role in helping the hand pose estimation. Similarly, a hand query point $\textbf{p}_h$ is also sent to object SDF decoder $\mathbb{SDF}_o$ and used to generate a cross-object query point feature $\textbf{f}_{ho}$.

\subsubsection{Feature Enhancement with Point-wise Attention}
\label{sec:feature_enhance}

As the pixel-aligned feature $\textbf{f}_{img}$ mainly contains local information, the local query point features $\textbf{f}_{h}$ and $\textbf{f}_{o}$ could be misled and thus make wrong predictions in the presence of severe occlusion. To address this problem, we propose to use an attention mechanism~\cite{vaswani2017attention,khan2022transformers} to exploit reliable dependencies in the global context. In contrast to existing approaches that either perform attention over 2D features~\cite{hampali2022keypoint} or over 3D mesh vertex features~\cite{tse2022collaborative, wang2023interacting}, our point-wise attention explores the global field information and the local image information with the aim of finding global 3D shape consistency between the sampled query points. Specifically, the extracted $N_h$ hand query point features $\{\textbf{f}_{h}^{i}\}_{i\in(0, N_h)}$ are sent into a hand attention module, which consists of six Multi-Head Self-Attention (MHSA) layers~\cite{vaswani2017attention, khan2022transformers}.

Meanwhile, to leverage object cues inside the cross-hand query point features $\{\textbf{f}_{oh}^{i}\}_{i\in(0, N_o)}$, we also send them to the MHSA layers $\mathbb{SA}$ to conduct cross attention with $\{\textbf{f}_{h}^{i}\}_{i\in(0, N_h)}$. The resulting enhanced hand point features are computed as
\begin{align}
\label{equ:feature_attention}
(\{\textbf{f}_{eh}^{i}\}_{i\in(0, N_h)}, *) = \mathbb{SA}(\{\textbf{f}_{h}^{i}\}_{i\in(0, N_h)}, \{\textbf{f}_{oh}^{i}\}_{i\in(0, N_o)}),
\end{align}
where $*$ denotes that we ignore the output from the $N_o$ cross-hand query tokens.
Analogously, the enhanced object point features $\{\textbf{f}_{eo}^{i}\}_{i\in(0, N_o)}$ can be obtained by processing object query point features $\{\textbf{f}_{o}^{i}\}_{i\in(0, N_o)}$ and cross-object query point features $\{\textbf{f}_{ho}^{i}\}_{i\in(0, N_h)}$ with an object attention module. 

\subsubsection{Point-wise Pose Regression} 
\label{sec:regressor}

With attention, we incorporate globally consistent information and cross-target cues into the hand point features $\{\textbf{f}_{eh}^{i}\}$ and object point features $\{\textbf{f}_{eo}^{i}\}$. Those points thus have enough global-local shape context information to regress hand-object poses. We apply asymmetric designs for hand and object pose estimation. Since the hand is non-rigid, flexible, and typically occluded when grasping an object, regressing the hand pose requires gathering richer information inside the $\{\textbf{f}_{eh}^{i}\}$. We hence follow \cite{hampali2022keypoint} to use Cross-Attention layers $\mathbb{CA}$ with the learned hand pose queries $\{\textbf{q}^{i}\}$. We supervise the learning of hand pose queries with MANO parameters \cite{romero2017embodied} to obtain both hand joints and a hand mesh. Sixteen hand pose queries regress 3-D MANO joint angles, and one more hand pose query regresses the 10-D mano shape parameters $\beta$. This can be expressed as
\begin{multline}
\label{equ:mano_regression}
(\{\bm{\theta}^{i} \in \mathbb{R}^3\}_{i\in(0, 16)}, \beta) = \\
\mathbb{CA}(\{\textbf{f}_{eh}^{i}\}_{i\in(0, N_h)}, (\{\textbf{q}^{i}\}_{i\in(0, 16)}, \textbf{q}^{16})).
\end{multline}

We use a smooth-L1 loss~\cite{ren2015faster} to supervise the learning of the MANO parameters, referred to as $\mathcal{L}_{\text{mano}}$. Similarly to \cite{hampali2022keypoint}, we also regress the intermediate hand pose objective to guide the final predictions. However, since our $\{\textbf{f}_{eh}^{i}\}$ already contains rich 3D information, we directly regress 3D hand joints instead of 2D joints as in \cite{hampali2022keypoint}. 
We use $\{\textbf{f}_{eh}^{i}\}$ as dense local regressors~\cite{li2019point, xiong2019a2j} to predict the offsets $\{\textbf{o}_h^{ij}\}$ from each hand query point $\textbf{p}_h^i$ to every pose joint as well as the prediction confidence. The corresponding loss is denoted as $\mathcal{L}_{\text{off}}$. Note that the design of the hand pose regressor is not identical. Our field-guided query points already include rich global-local shape context information and yield satisfactory pose estimation results with various regressors (see Sec.~\ref{sec:components} and Fig.~\ref{fig:pose_regressor}).

Compared with the hand, the object is more rigid. Therefore, we simply regress rotation vectors $\{\textbf{r}^i\}$ and translation vectors $\{\textbf{t}^i\}$ with all the enhanced object point features $\{\textbf{f}_{eo}^{i}\}$ and use a smooth-L1 loss~\cite{ren2015faster} $\mathcal{L}_{\text{obj}}$ to supervise them. During inference, we average the predictions from all the object points to obtain the final object translation and orientation.

\section{Experiments}

We first introduce the hand-object interaction benchmarks, describe implementation details and compare HOISDF with state-of-the-art (SOTA) methods. We finally detail  ablation results. 

\subsection{Datasets and Evaluation Metrics}
\label{sec:datasets}

We evaluate HOISDF on hand-object benchmarks: DexYCB~\cite{chao2021dexycb} and HO3Dv2~\cite{hampali2020honnotate} containing, respectively, 582K and 77K images of human interacting with YCB objects~\cite{calli2015benchmarking}. 
 
\textbf{DexYCB Dataset.} We use the default S0 train-test split defined by DexYCB~\cite{chao2021dexycb}. Some methods~\cite{liu2021semi, lin2023harmonious} use the full DexYCB dataset by flipping the left-hand images (denoted as DexYCB Full), while other methods~\cite{hasson2019learning, hasson2021towards, yang2022artiboost, chen2022alignsdf, chen2023gsdf, tse2022collaborative, wang2023interacting} select input frames in which the right hand and the object are in close interaction to ensure the physical contact (denoted as DexYCB). In general when we refer to DexYCB we mean this latter split. To broadly compare, we train HOISDF on both settings. Since most of the methods use the data only with the right hand, we conduct our ablations under the DexYCB split.

For hand pose estimation, we report Mean Joint Error (MJE) and Procrustes Aligned Mean Joint Error (PAMJE)~\cite{zimmermann2017learning}. We also report Mean Mesh Error (MME), area under the curve of the percentage of correct vertices (VAUC) the F-scores (F@5mm and F@15mm), and corresponding Procrustes Aligned version following~\cite{xu2023h2onet} to measure hand mesh reconstruction performance. For object 6D pose estimation, we report Object Center Error (OCE) following~\cite{chen2022alignsdf, chen2023gsdf}, Mean Corner error (MCE) following \cite{wang2023interacting}, and standard pose estimation average closest
point distance (ADD-S) following~\cite{hasson2019learning, hasson2021towards, wang2023interacting} to measure performance in center, corner, and vertex levels.

\textbf{HO3Dv2 Dataset.} We use the standard train-test splitting protocol and submit the test results to the official website to report performance. Since the HO3Dv2 is relatively small-scale, some methods~\cite{yang2022artiboost, wang2023interacting} render synthetic hand object images to enhance learning. Therefore, apart from training the model only with the original data in the HO3Dv2 training set, we also train another model (denoted with `*` in Table~\ref{tab:HO3D}) by including synthetic images. We follow the render pipeline of Wang et al.~\cite{wang2023interacting}. 

For hand pose estimation, we use the HO3Dv2 evaluation metrics to measure the performance: Mean Joint Error (MJE), Scale-Translation aligned Mean Joint Error (STMJE)~\cite{zimmermann2019freihand}, and Procrustes aligned Mean Joint Error (P-MJE)~\cite{zimmermann2017learning}. For object 6D pose estimation, we report mean Object Mesh Error (OME) and standard pose estimation average closest point distance (ADD-S) following \cite{hasson2019learning, hasson2021towards, wang2023interacting}.

 \subsection{Implementation and Training Details}
 \label{sec:details}
 
We adopt ResNet-50 as the U-Net backbone~\cite{ronneberger2015u,he2016deep}. All the point features: the image $\textbf{f}_{img}$, the hand $\textbf{f}_{eh}$, and object $\textbf{f}_{eo}$ are of size 256. We employ a transformer~\cite{vaswani2017attention} encoder as our point-wise attention module and a transformer decoder as our MANO regressor \cite{romero2017embodied}. We follow the standard practice \cite{hampali2022keypoint, lin2023harmonious, wang2023interacting} to train a unified model for all the objects in the dataset. The overall loss is a weighted sum of all individual loss functions, 
\begin{multline}
\label{equ:loss_all}
\mathcal{L}= \lambda_1\mathcal{L}_{img} + \lambda_2\mathcal{L}_{sdf} + \lambda_3\mathcal{L}_{\text{mano}} + \\ \lambda_4\mathcal{L}_{\text{off}} +  \lambda_5\mathcal{L}_{\text{obj}},
\end{multline}
where $\lambda_1$ to $\lambda_5$ are used to balance all the loss terms to the same scale. During training, the network parameters are optimized with Adam~\cite{kingma2014adam} with a mini-batch size of 32. The initial learning rate is 1e-4 and decays by 0.7 every 5 epochs. HOISDF typically converges to a satisfying result after about 40 epochs.

For query points sampling, during training, we sample $N_s=1000$ query points for 3D field learning. During inference, we empirically found that with a discretization size of $N_v=64$, sampling $N_v^2/n_h=600$ hand query points and $N_v^2/n_o=200$ object query points was enough for good performance.

\begin{table}[tp]
	\scriptsize
	\setlength{\tabcolsep}{1.8mm}
	\begin{center}
		\begin{tabular}{c|cc|cccc}   \toprule
        Metrics in [mm]  & MJE & PAMJE & OCE & MCE & ADD-S & Object\\
        \midrule
        Lin \emph{et al.} \cite{lin2021end} & 15.2 & 6.99 & - & - & - & No \\
        Spurr \emph{et al.} \cite{romero2017embodied} & 17.3 & 6.83 & - & - & - & No \\
        Liu \emph{et al.} \cite{liu2021semi} & 15.2 & 6.58 & - & - & - & Yes \\
        Park \emph{et al.} \cite{park2022handoccnet} & 14.0 & 5.80 & - & - & - & No \\
        Chen \emph{et al.} \cite{chen2022mobrecon} & 14.2 & 6.40 & - & - & - & No \\
        Xu \emph{et al.} \cite{xu2023h2onet} & 14.0 & 5.70 & - & - & - & No \\
        Lin \emph{et al.} \cite{lin2023harmonious} & 12.6 & 5.47 & 42.7 & 48.0 & 33.8 & Yes \\
        \midrule
        HOISDF (ours) & \textbf{10.1} & \textbf{5.13} & \textbf{27.6} & \textbf{35.8} & \textbf{18.6} & Yes \\
        \bottomrule
		\end{tabular}
	\end{center}
  \vspace{-.5cm}
	\caption{Quantitative comparison on the DexYCB dataset. Trained and tested on the DexYCB Full split. HOISDF reaches lower hand and object pose estimation errors. The metrics are represented in millimeters. The last column indicates whether a method performs the object 6D pose estimation.}
	\label{tab:DexYCB}
  \vspace{-.2cm}
\end{table}

\begin{table}[tp]
	\scriptsize
	\setlength{\tabcolsep}{1.8mm}
	\begin{center}
		\begin{tabular}{c|cc|cccc}   \toprule
        Metrics in [mm]  & MJE & PAMJE & OCE & MCE & ADD-S & Object\\
        \midrule
        Hasson \emph{et al.} \cite{hasson2019learning} & 17.6 & - & - & - & - & Yes \\
        Hasson \emph{et al.} \cite{hasson2021towards} & 18.8 & - & - & 52.5 & - & Yes \\
        Tze \emph{et al.} \cite{tse2022collaborative} & 15.3 & - & - & - & - & Yes \\
        Li \emph{et al.}  \cite{yang2022artiboost} & 12.8 & - & - & - & - & Yes \\
        Chen \emph{et al.} \cite{chen2022alignsdf} &  19.0 & - & 27.0 & - & - & Yes \\
        Chen \emph{et al.} \cite{chen2023gsdf} & 14.4 & - & 19.1 & - & - & Yes \\
        Wang \emph{et al.} \cite{wang2023interacting} & 12.7 & 6.86 & 27.3 & 32.6 & 15.9 & Yes \\
        Lin \emph{et al.} \cite{lin2023harmonious} & 11.9 & 5.81 & 39.8 & 45.7 & 31.9 & Yes \\
        \midrule
        HOISDF (ours) & \textbf{10.1} & \textbf{5.31} & \textbf{18.4} & \textbf{27.4} & \textbf{13.3} & Yes \\
        \bottomrule
		\end{tabular}
	\end{center}
 \vspace{-.5cm}
	\caption{Same as Table~\ref{tab:DexYCB}, but for DexYCB split, see Sec.~\ref{sec:datasets}.}
	\label{tab:DexYCB-S0}
  \vspace{-.5cm}
\end{table}

\begin{table*}[tp]
	\scriptsize
	\setlength{\tabcolsep}{2.5mm}
	\begin{center}
		\begin{tabular}{c|cccc|cccc|c}   \toprule
        Metrics  & MME$\downarrow$ & VAUC$\uparrow$ & F@5$\uparrow$ & F@15$\uparrow$ & PAMME$\downarrow$ & PAVAUC$\uparrow$ & PAF@5$\uparrow$ & PAF@15$\uparrow$ & Object \\
        \midrule
        Park \emph{et al.} \cite{park2022handoccnet} & 13.1 & 76.6 & 51.5 & 92.4 & 5.5 & 89.0 & 78.0 & 99.0 & No \\
        Chen \emph{et al.} \cite{chen2022mobrecon} & 13.1 & 76.1 & 50.8 & 92.1 & 5.6 & 88.9 & 78.5 & 98.8 & No \\
        Xu \emph{et al.} \cite{xu2023h2onet} & 13.0 & 76.2 & 51.3 & 92.1 & 5.5 & 89.1 & 80.1 & 99.0 & No \\
        Lin \emph{et al.} \cite{lin2023harmonious} & 11.6 & 77.6 & 53.0 & 93.3 & 5.2 & 89.6 & 79.8 & 99.2 & Yes \\
        \midrule
        HOISDF (ours) & \textbf{9.9} & \textbf{80.5} & \textbf{60.1} & \textbf{94.9} & \textbf{4.9} & \textbf{90.2} & \textbf{81.8} & \textbf{99.3} & Yes \\
        \bottomrule
		\end{tabular}
\vspace{-.5cm}
	\end{center}
	\caption{Quantitative comparison with hand mesh metrics on the DexYCB Full testset. MME and PAMME are in millimeters.}
	\label{tab:mesh}
\vspace{-0.5cm}
\end{table*}

\subsection{Comparisons with State-of-the-Art Methods}
\label{sec:sota_comparison}

\paragraph{Quantitative comparisons on DexYCB.} We evaluate HOISDF on the DexYCB test sets (Tables~\ref{tab:DexYCB}, \ref{tab:DexYCB-S0}, and Table~\ref{tab:perobj_dexycb} for per object results) and compare it with (SOTA) methods. Among the best models, \cite{wang2023interacting} is best at object estimation while \cite{lin2023harmonious} is best at hand pose estimation. However, HOISDF outperforms prior methods by a substantial margin for both hand and object metrics. Liu et al.\cite{liu2021semi} and Lin et al.\cite{lin2023harmonious} trained on the S0-DexYCB split. We train and test our HOISDF using the same split and observe a consistent improvement over them (Table~\ref{tab:DexYCB-S0}). It is also worth mentioning that HOISDF beats the methods that perform just hand pose estimation (e.g.,\cite{lin2021end, romero2017embodied, park2022handoccnet, xu2023h2onet}). Furthermore, we also compare HOISDF with SDF-based hand object interaction methods~\cite{chen2022alignsdf, chen2023gsdf}. As mentioned in Sec.~\ref{sec:field}, both of them use SDFs to regress the (output) hand meshes, while we use SDFs as intermediate representations and for field-guided inference. HOISDF significantly outperforms these methods. 

As HOISDF also predicts a MANO mesh, we compare it with the SOTA methods for hand mesh reconstruction performance on the DexYCB Full test set (Table~\ref{tab:mesh}). We observe consistent improvements with HOISDF. 

\paragraph{Quantitative comparisons on HO3Dv2.} As further evidence of the effectiveness of HOISDF, we also evaluate it on the HO3Dv2 dataset. Again, HOISDF consistently beats the current SOTA methods on almost all the hand and object metrics both with and without synthetic data (Table~\ref{tab:HO3D}, Table~\ref{tab:perobj_ho3d} for per object results). Lin et al.~\cite{lin2023harmonious} obtains slightly better performance with regard to PAMJE, but performs very poorly in the other metrics, while HOISDF is more balanced.

On both datasets, especially HO3Dv2 with fewer data, HOISDF yields a larger improvement on the metrics that exploit more global information (MJE, STMJE and object metrics). We attribute this advantage to the fact that SDFs, as intermediate representations, capture global information effectively to guide the subsequent pose estimations. We will first visualize query points (Fig. \ref{fig:intermediate_dexycb}) and then validate our design choices. 

\paragraph{Visualization of the learned SDFs.} We visualize the pose predictions and the intermediate hand-object query points on the DexYCB testset (Fig.~\ref{fig:intermediate_dexycb}). We can see that the remaining query points after the field-informed point sampling already reveal the general hand object shape (see Sec.~\ref{sec:sampling}).

\begin{figure}[!tp]
\begin{center}
\includegraphics[width=0.45\textwidth]{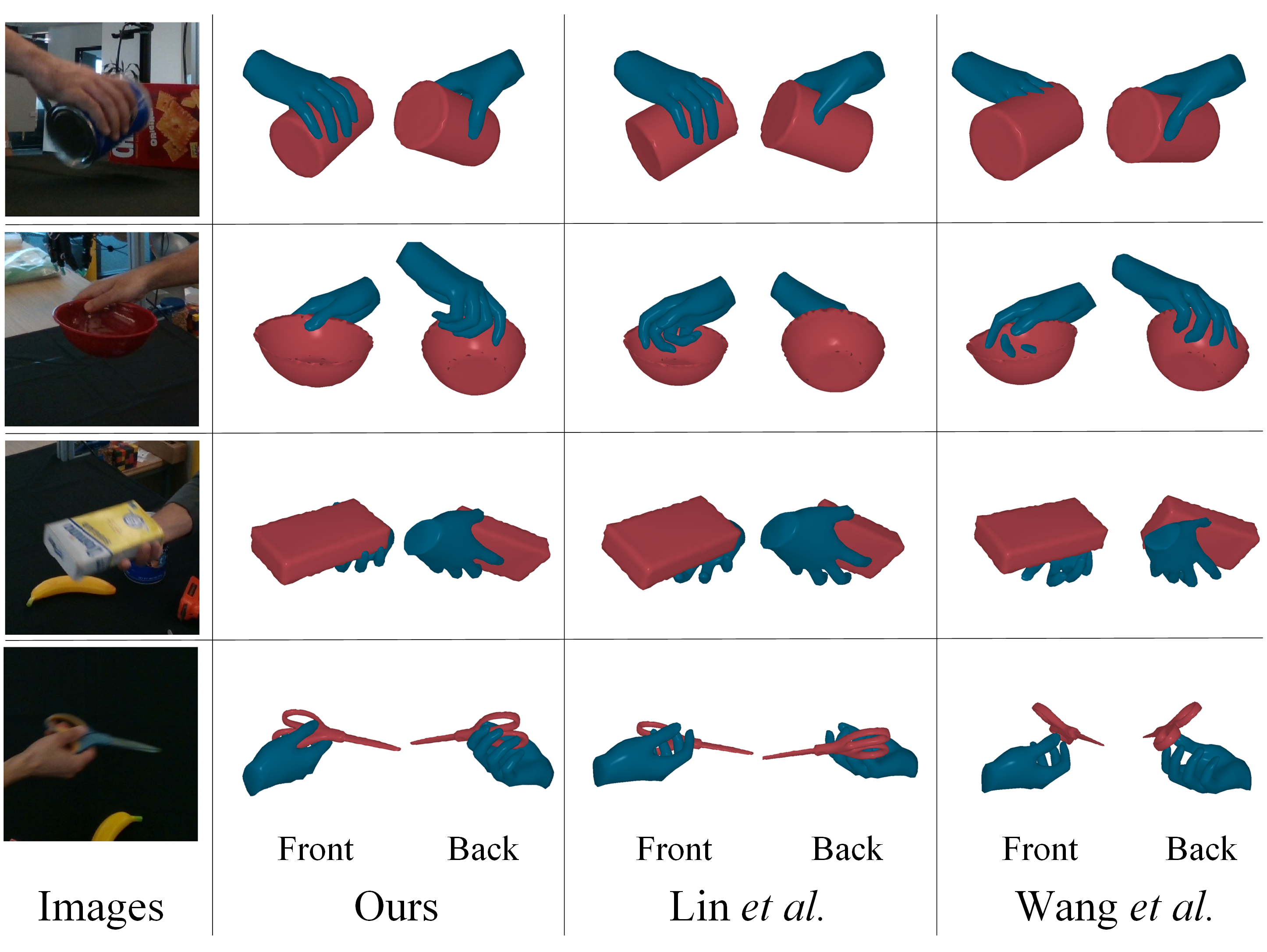}
\end{center} 
\vspace{-.5cm}
\caption{{\bf Qualitative comparisons between HOISDF and \cite{lin2023harmonious, wang2023interacting} on DexYCB testset.} HOISDF effectively uses robust global clues near the hand and object to deal well with various objects and severe occlusions.}
\label{fig:qualitative_dexycb}
\end{figure}

\paragraph{Qualitative comparisons.}
Next, we compared HOISDF qualitatively with two SOTA hand object pose estimation methods on the DexYCB test set (Fig.~\ref{fig:qualitative_dexycb}) and the HO3Dv2 test set (Fig.~\ref{fig:qualitative_ho3d}). We can see HOISDF outperforms \cite{lin2023harmonious, wang2023interacting} under various objects and different types of hand-object interactions.

\begin{table}[tp]
	\scriptsize
	\setlength{\tabcolsep}{2mm}
	\begin{center}
		\begin{tabular}{c|ccc|cc}   \toprule
        Metrics in [mm] & MJE & STMJE & PAMJE & OME & ADD-S \\
        \midrule
        Hasson \emph{et al.} \cite{hasson2019learning} & - & 31.8 & 11.0 & - & -  \\
        Hasson \emph{et al.} \cite{hasson2020leveraging} & - & 36.9 & 11.4 & 67.0 & 22.0  \\
        Hasson \emph{et al.} \cite{hasson2021towards} & - & 26.8 & 12.0 & 80.0 & 40.0  \\
        Liu \emph{et al.} \cite{liu2021semi} & - & 31.7 & 10.1 & - & - \\
        Hampali \emph{et al.} \cite{hampali2022keypoint} & 25.5 &  25.7 & 10.8 & 68.0 & 21.4 \\
        Lin \emph{et al.} \cite{lin2023harmonious} & 28.9 & 28.4 & \textbf{8.9} & 64.3 & 32.4 \\
        HOISDF (ours) & \textbf{23.6} & \textbf{22.8} & 9.6 & \textbf{48.5} & \textbf{17.8} \\
        \midrule
        Li \emph{et al.}*  \cite{yang2022artiboost} & 26.3 & 25.3 & 11.4 & - & -  \\
        Wang \emph{et al.}* \cite{wang2023interacting} & 22.2 & 23.8 & 10.1 & 45.5 & 20.8 \\
        HOISDF* (ours) & \textbf{19.0} & \textbf{18.3} & \textbf{9.2} & \textbf{35.5} & \textbf{14.4} \\
        \bottomrule
		\end{tabular}
  \vspace{-.5cm}
	\end{center}
	\caption{Quantitative comparison on the HO3Dv2 dataset. The metrics are represented in millimeters.`*` denotes models that were co-trained with synthetic data. }
	\label{tab:HO3D}
\vspace{-0.2cm}
\end{table}

\subsection{Ablation for Intermediate Representations}
\label{sec:ablation_representation}

Since using SDF as a global intermediate representation is the key component of HOISDF, we analyze the role of the SDF here, comparing it with other intermediate representations, and analyzing the query points.

\begin{table}[tp]
	\scriptsize
	\setlength{\tabcolsep}{2.5mm}
	\begin{center}
		\begin{tabular}{cccccc}   \toprule
        Metrics in [mm]  & MJE & PAMJE & OCE & MCE & ADD-S \\
        \midrule
        2D Keypoint & 14.9 & 7.13 & 34.2 & 45.3 & 22.9 \\
        2D Segmentation & 14.1 & 6.88 & 31.3 & 43.1 & 21.0 \\
        3D Vertices & 12.7 & 6.57 & 24.1 & 35.3 & 16.5 \\
        \midrule
        3D SDFs (ours)  & \textbf{10.1} & \textbf{5.31} & \textbf{18.4} & \textbf{27.4} & \textbf{13.3} \\
        \bottomrule
		\end{tabular}
	\end{center}
 \vspace{-.5cm}
	\caption{Comparison between different intermediate representations on DexYCB testset. The SDF-based intermediate representation outperforms other representations because it encodes 3D shape information, is direct to regress, and has less joint cumulative error.}
	\label{tab:representation}
\vspace{-0.4cm}
\end{table}

\paragraph{Comparison of different intermediate representations.} Here, to elucidate the role of the SDF, we build several baselines that use different intermediate representations while trying to keep the remaining model components (e.g., image backbones, feature dimensions, pose regressors, etc.) the same as in our model. We replace the 3D field learning module (Sec.~\ref{sec:field}) with 2D keypoint learning, 2D segmentation learning, and 3D mesh learning (see Supp. Mat.~\ref{sec:interm-ablation}). We found that utilizing intermediate 2D representations is much worse, and that 3D vertices are also significantly less powerful than SDFs (Table~\ref{tab:representation}). Next, we provide further evidence for the effectiveness of using SDF as an intermediate representation by analyzing the sampled query points during inference.

\begin{table}[tp]
	\scriptsize
	\setlength{\tabcolsep}{3mm}
	\begin{center}
		\begin{tabular}{cccccc}   \toprule
        Metrics in [mm]  & Mean & MCP & PIP & DIP & Tip \\
        \midrule
        Wang \emph{et al.} \cite{wang2023interacting} &  7.67 & 7.63 & 6.36 & 6.29 & 10.4 \\
        HOISDF (ours) &  \textbf{6.16} &  \textbf{6.02} & \textbf{5.27} & \textbf{5.40} & \textbf{7.95}\\
        \bottomrule
		\end{tabular}
	\end{center}
  \vspace{-.5cm}
	\caption{Sampled point distributions. Using SDF as global guidance for point sampling gathers intermediate query points closer to the GT pose joints. MCP, PIP, DIP, and Tip are different finger parts.}
	\label{tab:point_sampling}
  \vspace{-.4cm}
\end{table}

\paragraph{Analysis of the sampled points.} We argued that the SDF representation better captures global shape information across the capture volume (Figure~\ref{fig:fig1}). We analyze the point distributions of HOISDF's hand query points sampled using our proposed point sampling strategy and the intermediate hand mesh vertices extracted by the initial stage of Wang et al.~\cite{wang2023interacting}. Indeed, our model samples closer points to the hand joints, particularly for the most challenging finger joints like the finger tips (Table~\ref{tab:point_sampling}).

\subsection{Ablations for the Field-Guided Pose Regression Module}
\label{sec:components}

The field-guided pose regression module is the other key component to let HOISDF effectively leverage the SDF information. To verify that, we conduct ablations for different parts. Firstly, we showed that our field-guided sampling method is efficient and robust by comparing it with other sampling ways (Table~\ref{tab:sampling_way}). Secondly, we assessed the role of the point feature augmentation method by comparing it with different variations; altering various parts gracefully reduced the performance (Table~\ref{tab:fea_aug}). Next, the mutual hand-object feature enhancement method proposed in Sec.\ref{sec:cross_interaction} is also proven to be effective by removing the cross attention or replacing with other non-augmented features (Table~\ref{tab:fea_cross}). Finally, we show that HOISDF is robust to changes in regression targets (Table~\ref{tab:regressor}) since our hand/object query points already capture enough global-local context with our field-guided module. Overall, these ablations validate our design choices.

\begin{table}[tp]
	\scriptsize
	\setlength{\tabcolsep}{2mm}
	\begin{center}
		\begin{tabular}{cccccc}   \toprule
        Metrics in [mm]  & MJE & PAMJE & OCE & MCE & ADD-S \\
        \midrule
        Random & 25.8 & 13.5 & 48.4 & 53.7 & 29.6 \\
        Signed distance & 13.3 & 6.58 & 19.7 & 30.7 & 15.9 \\
        Field gradient & \textbf{10.1} & \textbf{5.29} & 18.5 & 27.7 & 13.5 \\
        \midrule
        Absolute distance (ours)  & \textbf{10.1} & 5.31 & \textbf{18.4} & \textbf{27.4} & \textbf{13.3} \\
        \bottomrule
		\end{tabular}
	\end{center}
  \vspace{-.5cm}
	\caption{Comparison between different sampling strategies on DexYCB testset. Our field-informed point sampling can achieve the best performance. See Supp. Mat.~\ref{sec:fgp-ablations} for details on the alternative sampling strategies.}
	\label{tab:sampling_way}
\vspace{-0.3cm}
\end{table}

\begin{table}[tp]
	\scriptsize
	\setlength{\tabcolsep}{2mm}
	\begin{center}
		\begin{tabular}{cccccc}   \toprule
        Metrics in [mm]  & MJE & PAMJE & OCE & MCE & ADD-S \\
        \midrule
        w/o SDF augmentation & 11.5 & 6.05 & 23.6 & 31.2 & 15.7 \\
        w density concatenation & 11.0 & 5.71 & 22.7 & 30.5 & 15.3 \\
        w distance concatenation & 11.5 & 6.07 & 23.3 & 30.9 & 15.6 \\
        \midrule
        w SDF augmentation & \textbf{10.8} & \textbf{5.68} & \textbf{22.2} & \textbf{30.0} & \textbf{15.1} \\
        \bottomrule
		\end{tabular}
	\end{center}
  \vspace{-.5cm}
	\caption{Effects of field-based point feature augmentation on the DexYCB test set. Our SDF feature augmentation best enhances features for the subsequent pose estimations. See Supp. Mat.~\ref{sec:fgp-ablations} for details on the alternative augmentations. }
	\label{tab:fea_aug}
 \vspace{-.2cm}
\end{table}

\begin{table}[tp]
	\scriptsize
	\setlength{\tabcolsep}{1.2mm}
	\begin{center}
		\begin{tabular}{cccccc}   \toprule
        Metrics in [mm]  & MJE & PAMJE & OCE & MCE & ADD-S \\
        \midrule        
        w/o cross feature enhancement & 10.8 & 5.68 & 22.2 & 30.0 & 15.1 \\
        w cross image feature & 11.1 & 5.74 & 20.2 & 28.6 & 14.2 \\
        w cross target feature & 11.3 & 5.81 & 23.7 & 31.8 & 15.9 \\
        \midrule
        Cross feature enhancement (ours) & \textbf{10.1} & \textbf{5.31} & \textbf{18.4} & \textbf{27.4} & \textbf{13.3} \\
        \bottomrule
		\end{tabular}
	\end{center}
  \vspace{-.5cm}
	\caption{Effects of hand-object feature enhancement on the DexYCB testset. HOISDF's cross feature enhancement gave the best results. See Supp. Mat.~\ref{sec:fgp-ablations} for details on the alternative feature computations.}
	\label{tab:fea_cross}
 \vspace{-.2cm}
\end{table}

\begin{table}[tp]
	\scriptsize
	\setlength{\tabcolsep}{3mm}
	\begin{center}
		\begin{tabular}{ccc}   \toprule
        Metrics in [mm]  & MJE & PAMJE \\
        \midrule
        w/o intermediate joint regression & 10.4 & 5.49 \\
        w/o MANO regression & 10.5 & 5.65 \\
        w MANO shape \& inverse kinematics & \textbf{10.0} & 5.35 \\
        \midrule
        MANO regression (Ours) & 10.1 & \textbf{5.31} \\
        \bottomrule
		\end{tabular}
	\end{center}
  \vspace{-.5cm}
	\caption{Robustness to different pose regressors on the DexYCB testset. Benefiting from the rich global-local context information inside the enhanced features, HOISDF can obtain great performance even with simple pose regression targets. See Supp. Mat.~\ref{sec:fgp-ablations} for details on the alternative regression targets.}
	\label{tab:regressor}
\end{table}

\section{Conclusion}
\label{sec:conclusion}

We proposed a novel 3D hand-object pose estimation algorithm that takes advantage of jointly learned signed distance fields. It achieves strong results and inference is fast (see Sup. Mat.~\ref{sec:speed}) We believe this paradigm could also be applied to other pose estimation problems, e.g.,~\cite{chen2019overview,billard2019trends,ren2020review,mathis2020deep,chiappa2024acquiring}.

\section*{Acknowledgments}

The work was funded by EPFL and Microsoft Swiss Joint Research Center (H.Q., A.M.). H.Q. acknowledges support from a Boehringer Ingelheim Fonds PhD stipend. We are grateful to the members of the Mathis Group and in particular Niels Poulsen for comments on an earlier version of this manuscript. We also sincerely thank Rong Wang, Wei Mao and Hongdong Li for sharing the hand-object rendering pipeline~\cite{wang2023interacting}.


\section*{Supplementary materials}

\appendix

\renewcommand\thefigure{F\arabic{figure}}

\setcounter{figure}{0}

\renewcommand\thetable{T\arabic{table}}
\setcounter{table}{0}

We first provide additional details on the architecture design of HOISDF with respect to the image feature extraction and hand pose regression. Then, we provide additional details for the ablation experiments. Finally, we conduct additional experiments to assess the effectiveness of HOISDF.

\setcounter{section}{0}

\section{Architecture details}
\label{sec:architecture-details}

\subsection{Image Feature Extraction}

Here, we detail the regressed objectives and the corresponding losses for the image backbone mentioned in Sec.~\ref{sec:global_sdf}. Following standard practice~\cite{hampali2022keypoint, wang2023interacting, lin2023harmonious}, we regress 2D heatmaps and hand/object segmentation masks as additional 2D predictions. Specifically, for simplicity, we regress a single-channel 2D hand keypoints heatmap $\textbf{H}_h$~\cite{hampali2022keypoint}. To obtain the ground-truth heatmap $\textbf{H}^*_h$, we convolve all the 2D joint locations with a 2D Gaussian kernel and sum them in the same channel. Furthermore, we regress the hand and object segmentation maps ($\textbf{H}_s$ and $\textbf{O}_s$) as two additional channels. To learn $\textbf{H}_h$, $\textbf{H}_s$, and $\textbf{O}_s$, we minimize the loss
\begin{align}
\label{equ:loss_img}
\mathcal{L}_{img}={\|\textbf{H}^*_h - \textbf{H}_h\|} + \mathcal{CE}(\textbf{H}_s, \textbf{H}_s^*) + \mathcal{CE}(\textbf{O}_s, \textbf{O}_s^*),
\end{align}
where $\mathcal{CE}$ represents the cross-entropy loss, and $\textbf{H}^*_s$ and $\textbf{O}_s^*$ are obtained by rendering the ground-truth 3D hand and object meshes.

\subsection{Hand pose regression}

In Sec.~\ref{sec:pose_attention}, we show that the field-guided pose regression module uses the point-wise features augmented by the field information to predict the hand object poses. Here, we give more details about the hand pose estimation component.

\begin{figure}[tp]
\begin{center}
\includegraphics[width=0.42\textwidth]{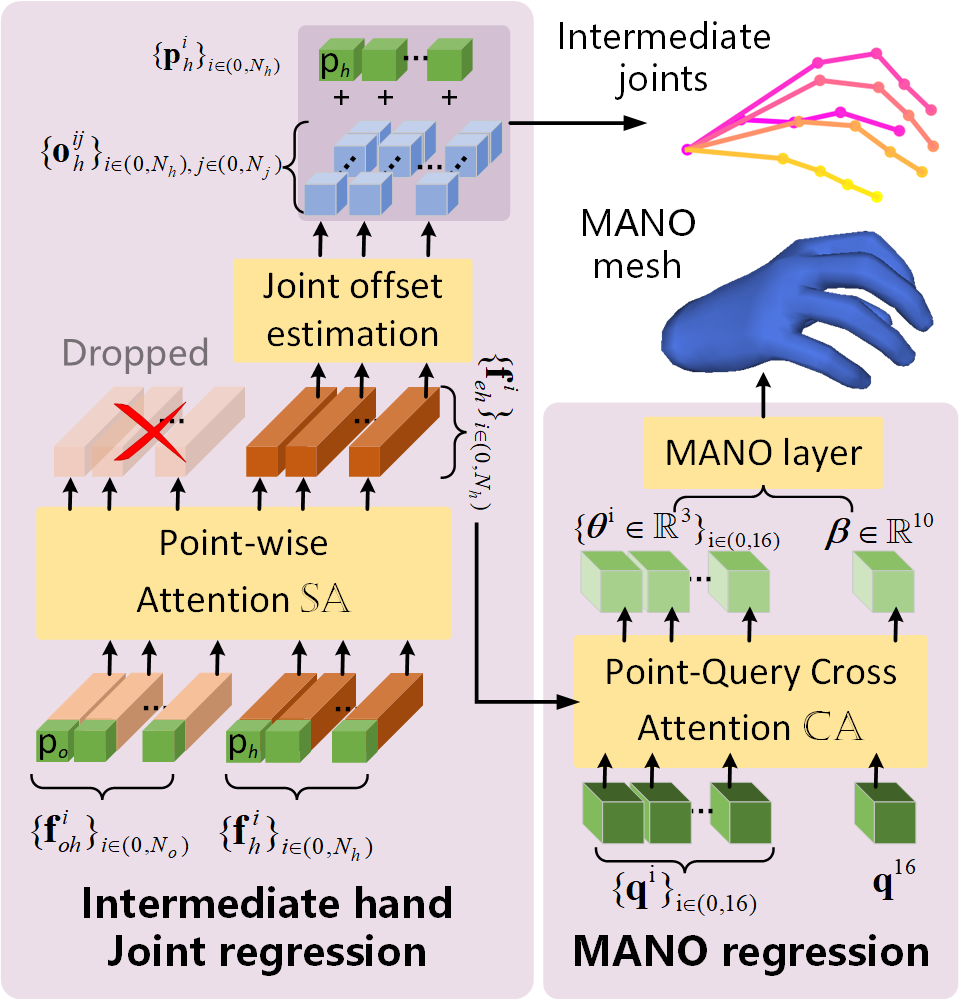}
\end{center} 
\vspace{-.5cm}
\caption{{\bf Details of hand pose regression of HOISDF.}}
\label{fig:pose_regressor}
\vspace{-0.6cm}
\end{figure}

As is shown in Figure \ref{fig:pose_regressor}, with the set of hand query point features $\{\textbf{f}_{h}^{i}\}_{i\in(0, N_h)}$ illustrated in Sec.~\ref{sec:fea_reg} and the set of cross-hand query point features $\{\textbf{f}_{oh}^{i}\}_{i\in(0, N_o)}$ illustrated in Sec.~\ref{sec:cross_interaction}, we conduct point-wise attention $\mathbb{SA}$ between all the point features. The resulting features from $\{\textbf{f}_{h}^{i}\}_{i\in(0, N_h)}$ are denoted as enhanced hand point features $\{\textbf{f}_{eh}^{i}\}_{i\in(0, N_h)}$, while the resulting features from $\{\textbf{f}_{oh}^{i}\}_{i\in(0, N_o)}$ are dropped since the object clues are already passed to $\{\textbf{f}_{eh}^{i}\}_{i\in(0, N_h)}$ through $\mathbb{SA}$ (illustrated in Sec.~\ref{sec:feature_enhance}).

We then conduct cross-attention between $\{\textbf{f}_{eh}^{i}\}_{i\in(0, N_h)}$ and learnable queries $\{\textbf{q}^{i}\}_{i\in(0, 17)}$, where the last query is used to regress MANO shape parameters $\beta\in \mathbb{R}^10$ and the rest queries are used to regress MANO pose parameters $\{\bm{\theta}^{i} \in \mathbb{R}^3\}_{i\in(0, 16)}$ (Eq.~\ref{equ:mano_regression}).

Meanwhile, similarly to Hampali et al.~\cite{hampali2022keypoint}, we also regress the intermediate hand pose objective to guide the final predictions. However, since the features $\{\textbf{f}_{eh}^{i}\}$ already contains rich 3D information, we directly regress 3D hand joints instead of 2D joints as in Hampali et al.\cite{hampali2022keypoint} and use the query points as dense local regressors~\cite{li2019point, xiong2019a2j}. Specifically, we use a joint offset regression head to predict the offsets $\{\textbf{o}_h^{ij}\}_{i\in(0, N_h),j\in(0,N_j)}$ from a hand query point $\textbf{p}_h^i$ to all the pose joints $\{{\textbf{h}^*}^j_p\}$, where $j$ represents the pose joint index and $N_j$ is the number of the hand pose joints. We use a smooth-L1 loss~\cite{ren2015faster} to supervise the learning of the offsets. However, if $\textbf{p}_h^i$ is far away from the pose joint ${\textbf{h}^*}^j_p$, the predicted $\textbf{o}_h^{ij}$ could be inaccurate. Therefore, instead of regressing all the joint offsets, we use a joint visibility term to determine if $\textbf{p}_h^i$ is close to ${\textbf{h}^*}^j$. We empirically set the joint class ${\textbf{v}^*}_h^{ij}$ to one if the distance between $\textbf{p}_h^i$ and $\textbf{h}^j$ is smaller than 4 cm, and to zero otherwise.
The joint visibility information is not accessible during inference. Therefore, we introduce a joint classification head to learn it. To train it, we minimize the cross entropy loss $\mathcal{CE}$ between the predicted joint visibility $\textbf{v}_h^{ij}$ and the ground truth ${\textbf{v}^*}_h^{ij}$. During inference, the predicted joint visibility $\{\textbf{v}_h^{ij}\}_i$ is sent to the SoftMax function~\cite{bridle1990probabilistic} to weigh the joint predictions. Altogether, this yields the training loss
\begin{multline}
\label{loss_handpose}
\mathcal{L}_{\text{off}}=\sum_{i}^{N_h} \sum_{j}^{N_j} SmoothL1(\textbf{p}_h^i+\textbf{o}_h^{ij}, {\textbf{h}^*}^j_p) \cdot {\textbf{v}^*}_h^{ij} \\
+ \mathcal{CE}(\textbf{v}_h^{ij}, {\textbf{v}^*}_h^{ij}).
\end{multline}

\section{Ablation Details}

\subsection{Comparison of different intermediate representations}

\label{sec:interm-ablation}

As discussed in Sec.~\ref{sec:ablation_representation}, we replace the 3D field learning module (Sec.~\ref{sec:field}) with 2D keypoint learning, 2D segmentation learning, and 3D mesh learning. Here, we give more details for the model designs of using other intermediate representations. Specifically, for 2D keypoint learning, we borrow the model design of Hampali et al.~\cite{hampali2022keypoint} to regress identity-aware but part-agnostic keypoints in the intermediate step to serve as query points. For 2D segmentation learning, we use the pixel locations with segmentation scores larger than 0.3 as query points. The keypoint confidence and the segmentation score are used to multiply with the query point features separately to mimic our feature regularization (Sec.~\ref{sec:fea_reg}) in the above two baselines. For 3D mesh learning, we follow Tse et al.~\cite{tse2022collaborative} to regress MANO parameters in the intermediate stage and use the MANO hand vertices to serve as hand query points. Meanwhile, we regress the object rotation and translation in the intermediate stage to obtain object vertices as object query points. 

We find that the SDF representation outperforms the 2D representations by a large margin, especially in MJE and object metrics that exploit more global information (Table~\ref{tab:representation}). We attribute this to the 2D intermediate representations gathering less 3D shape information in the initial step. Furthermore, we observe that using 3D vertices as intermediate representation performs better than 2D representations (Table~\ref{tab:representation}). This supports our claim that implicit 3D shape representations are better than explicit 3D meshes.

\subsection{Comparison to other SDF-based methods}

The key difference between HOISDF and other SDF-based methods~\cite{chen2022alignsdf, chen2023gsdf, ye2022s} is the role of the SDF module. Previous methods rely on the SDF module to reconstruct fine-grained hand-object surfaces. Predicting the SDF is the endpoint of the models. The resulting SDF values are used to generate meshes directly. By contrast, HOISDF shows that SDFs are a great intermediate representation for hand-object pose estimation (Table \ref{tab:representation}). The extracted SDF values are sent to the field-guided pose regression module to provide 3D global shape information for hand-object pose estimation. In comparison, we obtained better pose estimates than previous SOTA SDF methods \cite{chen2022alignsdf, chen2023gsdf} (Table \ref{tab:DexYCB-S0}).

\begin{figure}[tp]
\begin{center}
\includegraphics[width=0.47\textwidth]{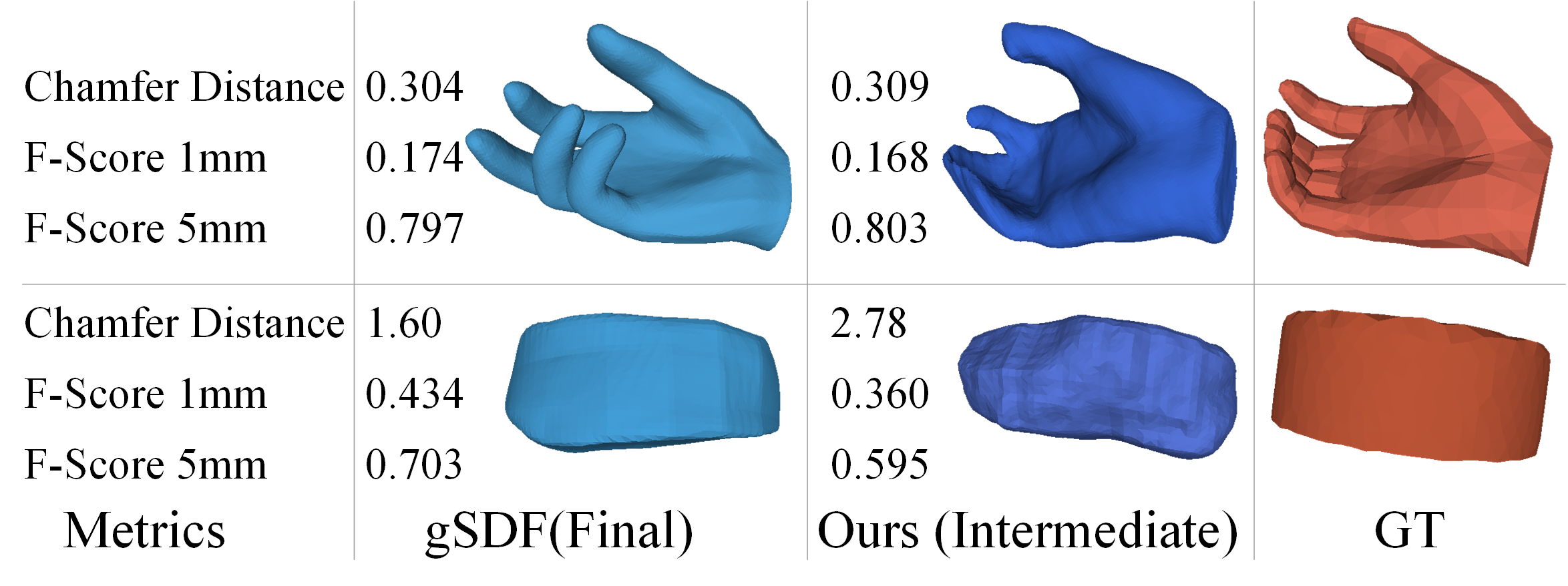}
\end{center} 
\vspace{-.6cm}
\caption{{\bf Comparisons between HOISDF's intermediate results and gSDF's \cite{chen2023gsdf} final results on DexYCB testset.} The SDF module in HOISDF cares more about global plausibility, while the one in gSDF cares more about fine-grained surface reconstruction.}
\label{fig:sdfcomparison}
\end{figure}

Due to different roles, the design choices of the SDF module in HOISDF and other SDF methods thus differ. To improve the quality of the reconstructed surfaces, previous methods \cite{chen2022alignsdf, chen2023gsdf, ye2022s} add intermediate pose regression modules. {\it The generated hand-object poses are used to pre-align the local parts with the canonical space. The SDF module can thus focus on fine-grained details without being disturbed by hand-object poses.} However, we aim to let the SDF module encode global pose information to guide the subsequent pose regression. We have evidence that adding a pose regression module before will convey unreliable pose information to the input of the SDF module and will pollute the global information captured by the SDF module (e.g., the little finger of gSDF's hand mesh in Figure \ref{fig:sdfcomparison}). Meanwhile, additional pose regression and canonicalization steps would also decrease the running speed of HOISDF and make the module unable to be end-to-end trained \cite{chen2023gsdf, ye2022s}.

To support our design choices, we directly use the intermediate SDF module to reconstruct hand-object meshes and compare them with gSDF's \cite{chen2023gsdf} final outputs (Figure \ref{fig:sdfcomparison})). Note that HOISDF also yields 3D hand and object meshes in the final outputs and obtains SOTA results (Table~\ref{tab:mesh} and Figure~\ref{fig:qualitative_dexycb}). Regarding our intermediate SDF module, we expect to have worse results since mesh reconstruction is not the goal of our SDF module. Surprisingly, however, it performs similarly to gSDF on hand metrics (Fig. \ref{fig:sdfcomparison}). We attribute this to the fact that our SDF module captures better global shape information. Therefore, even though the mesh reconstruction quality is lower, the overall distance to the GT hand mesh is acceptable. In comparison, the poses of the meshes produced by gSDF are influenced by its pose regression module and might yield large pose errors. As expected, our intermediate SDF module performs worse than gSDF on object metrics because of worse surface reconstruction. However, the general pose of our intermediate object reconstruction remains satisfactory. Note that gSDF is trained for 1600 epochs, while HOISDF is only trained for 40. We also replace our SDF module with gSDF initialized by their trained weights. The results (MJE: 11.2, PAMJE: 5.83, OCE: 19.6, MCE: 29.4, ADD-S: 14.3) show that despite more computational complexity, gSDF is less effective as an intermediate module.

\subsection{Ablations for the Field-guided Pose Regression Module}
\label{sec:fgp-ablations}
As discussed in Sec.~\ref{sec:components}, we verify the effectiveness of the components in our field-guided pose regression module by comparing each component with multiple variants. Here, we show the detailed designs of the variants.

\textbf{Effectiveness of the field-informed point sampling.} As discussed in Sec.~\ref{sec:sampling}, we sample query points close to the hand/object surfaces for the subsequent pose estimation. During inference, we sample query points with the smallest absolute distances to achieve the same goal. Here, we compare to three alternative point sampling strategies. The first one is to sample query points randomly in the 3D spaces. The second one is to sample query points inside the hand object meshes and sample points with the smallest signed distances during inference. The final one still samples points close to the hand-object surfaces. However, during the inference, we follow Zhou et al.~\cite{zhou2022learning} to compute the gradient of the SDF module according to a certain sampled query point. Then we multiply the gradient with the signed distance and use them as an offset to move the original sampled query point. This moves the query point even closer to the surfaces. Random sampling and signed distance sampling perform much worse than our absolute distance sampling, because the sampled points cannot reflect the general shapes of the hand and object and query irrelevant image features that will harm the pose estimation (Table~\ref{tab:sampling_way}). Applying field gradient to obtain the query points has almost the same performance as ours. However, computing the gradients for all the query points takes much more time compared to directly sampling points based on absolute distances. Therefore, in comparison our sampling strategy is the most efficient one.

\begin{figure}[!tp]
\begin{center}
\includegraphics[width=0.45\textwidth]{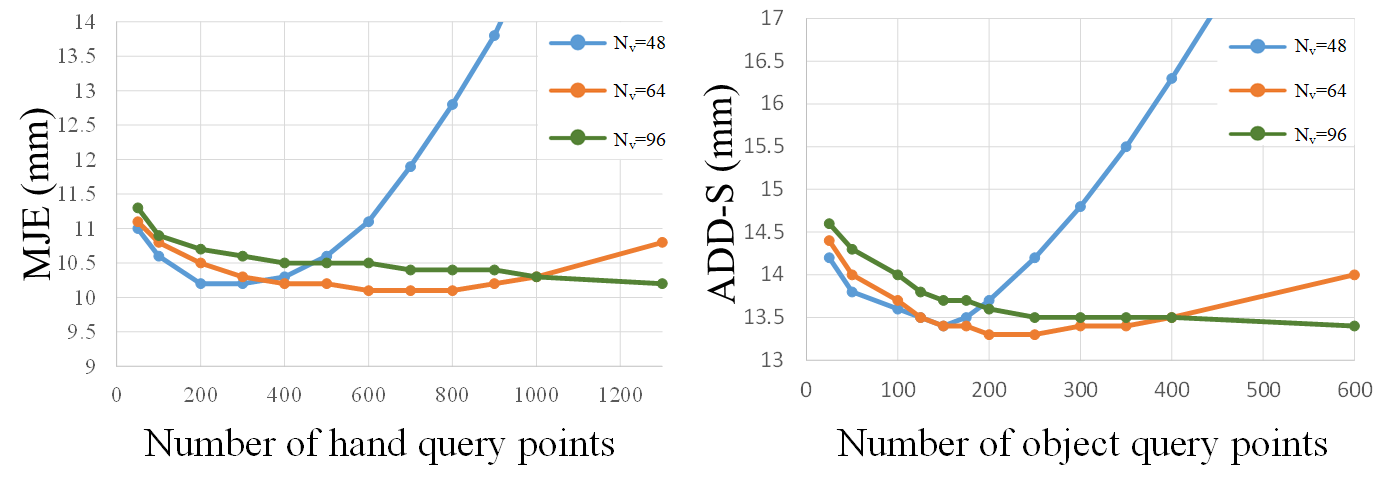}
\end{center} 
\vspace{-.5cm}
\caption{{\bf Hand object performance curve according to the numbers of sampled query points on DexYCB testset.} HOISDF is robust with a wide range of sampled query points under different discretization sizes.}
\label{fig:sampled_points}
\vspace{-0.5cm}
\end{figure}

\textbf{Effectiveness of field-based point feature augmentation.} As described in Sec.~\ref{sec:fea_reg}, we convert the point signed distance into a volume density and then multiply it with the point image feature to augment the feature. Since the cross hand object interaction (Sec.~\ref{sec:cross_interaction}) also uses the feature augmentation and will influence the performance, we remove the cross field attention and implement three variants to verify the effectiveness of the feature augmentation (Table \ref{tab:fea_aug}). Removing the SDF feature augmentation (\emph{w/o SDF regularization}), concatenating rather than multiplying the volume density with the image feature (\emph{w density concatenation}), and concatenating the distance value with the image feature (\emph{w distance concatenation}). Removing the SDF regularization yields an accuracy drop. Directly concatenating the distance values makes the model struggle to extract useful information. Directly concatenating the density value boosts the performance compared to \emph{w/o SDF regularization}. However, since it only has one dimension, it is hard to influence the whole feature representation.

\textbf{Effectiveness of hand-object feature enhancement.}
As discussed in Sec.~\ref{sec:cross_interaction}, we augment the object query point features with the cross-hand signed distances. The resulting cross-hand query point features are then used to conduct cross-attention with the original hand query point features to enhance the hand feature representation (Eqn.~\ref{equ:feature_attention}). Here, we conduct ablations to verify the effectiveness of our hand-object feature enhancement with three variants (Table \ref{tab:fea_cross}): Removing the cross feature enhancement completely (denoted as w/o cross feature enhancement), cross attention with cross target image features $f_{img}$ without feature augmentation (denoted as w cross image feature), cross attention with cross target features $f_h$ and $f_o$ (denoted as {\it w cross target feature}). Compared to {\it w/o cross feature enhancement}, both hand and object benefit from the cross target cues and improve the pose estimation performance. The variant {\it W cross image feature} only obtains very few improvements for the object pose estimation while has a side influence on the hand pose estimation. The object usually takes a larger space than the hand in the image. The various object features from different pixel locations will mislead the hand pose estimation without the guidance of the cross-hand signed distances. {\it W cross target feature} obtains the worst results for both hand and object pose estimations since the features are still augmented with the original signed distances instead of the cross-target signed distances, which are not helpful in transferring clues to the other target.

\begin{table}[tp]
	\scriptsize
	\begin{center}
		\centering
    \setlength{\tabcolsep}{1.5mm}
    \begin{tabular}{cccccccc}
			\toprule
			\textbf{Methods} & \multicolumn{3}{c}{HOISDF (ours)} && \multicolumn{3}{c}{Wang \emph{et al.} \cite{wang2023interacting}} \\
			\cline{2-4}\cline{6-8}
			\textbf{Metrics in [mm]} &   OCE & MCE & ADD-S  &&  OCE & MCE & ADD-S  \\
			\midrule
			002\_master\_chef\_can  & \textbf{15.9} & \textbf{20.2} & \textbf{10.2} && 21.8 & 25.5 & 12.8 \\
            003\_cracker\_box  & \textbf{29.4} & 40.2 & 18.5 && 33.3 & \textbf{37.8} & \textbf{17.8} \\
            004\_sugar\_box  & \textbf{17.1} & \textbf{29.7} & \textbf{14.2} && 24.6 & 32.3 & 14.7 \\
            005\_tomato\_soup\_can  & \textbf{17.9} & \textbf{20.8} & \textbf{10.3} && 29.4 & 31.7 & 15.0 \\
            006\_mustard\_bottle  & \textbf{13.6} & \textbf{18.1} & \textbf{9.1} && 20.4 & 24.5 & 11.1 \\
            007\_tuna\_fish\_can  & \textbf{15.4} & \textbf{17.3} & \textbf{8.9} && 23.6 & 24.5 & 12.5 \\
            008\_pudding\_box  & \textbf{13.3} & \textbf{19.5} & \textbf{9.5} && 21.0 & 24.5 & 12.1 \\
            009\_gelatin\_box  & \textbf{14.8} & \textbf{20.8} & \textbf{9.8} && 25.4 & 28.3 & 13.9 \\
            010\_potted\_meat\_can  & \textbf{13.9} & \textbf{19.8} & \textbf{10.5} && 24.7 & 26.7 & 12.4 \\
            011\_banana  & \textbf{19.5} & \textbf{41.7} & \textbf{20.6} && 28.1 & 42.2 & 21.0 \\
            019\_pitcher\_base  & \textbf{27.9} & \textbf{39.5} & \textbf{18.8} && 37.3 & 44.4 & 21.5 \\
            021\_bleach\_cleanser  & \textbf{19.0} & 40.9 & 18.6 && 34.4 & \textbf{39.7} & \textbf{17.8} \\
            024\_bowl  & \textbf{17.7} & \textbf{21.5} & \textbf{12.0} && 28.5 & 30.2 & 16.1 \\
            025\_mug  & \textbf{16.5} & \textbf{17.9} & \textbf{9.5} && 27.1 & 27.3 & 12.3 \\
            035\_power\_drill  & \textbf{20.5} & 31.2 & 16.1 && 26.8 & \textbf{30.8} & \textbf{14.5} \\
            036\_wood\_block  & \textbf{27.9} & \textbf{35.3} & \textbf{17.1} && 35.8 & 46.4 & 21.7 \\
            037\_scissors  & \textbf{25.4} & 49.0 & 21.3 && 33.5 & \textbf{47.8} & \textbf{22.8} \\
            040\_large\_marker  & \textbf{14.9} & \textbf{24.2} & \textbf{12.9} && 25.1 & 31.8 & 18.3 \\
            052\_extra\_large\_clamp  & \textbf{23.7} & 48.3 & \textbf{22.4} && 31.2 & \textbf{45.8} & 22.7 \\
            061\_foam\_brick  & \textbf{13.7} & \textbf{16.3} & \textbf{8.0} && 24.3 & 25.1 & 11.4 \\
			\midrule 
            Mean & \textbf{18.4} & \textbf{27.4} & \textbf{13.3} && 27.3 & 32.6 & 15.9 \\
            \bottomrule
		\end{tabular}
	\end{center}
    \vspace{-0.5cm}
	\caption{\textbf{Per-object performance on DexYCB testset.} Our HOISDF can outperform Wang \emph{et al.} \cite{wang2023interacting} for most of the objects, demonstrating HOISDF is robust to various objects. }
	\label{tab:perobj_dexycb}
 \vspace{-0.3cm}
\end{table}

\textbf{Robustness with various pose regression components.}
As mentioned in Sec.~\ref{sec:regressor}, we use learnable queries to conduct cross-attention with enhanced hand query point features $\{\textbf{f}_{eh}^{i}\}$ and regress the MANO parameters. Note, however, that the strong hand pose estimation performance is mainly because of the field-based feature enhancement rather than the design of the hand pose regressor. To verify that, we also implement three other hand pose regressors (Table \ref{tab:regressor}). The first one removes the intermediate hand joint regression. The second one removes the cross-attention layer and directly uses the intermediate hand joints as the final result. The last one only uses the cross-attention layer to regress the MANO shape parameters. The MANO pose parameters are inferred from the intermediate hand joints using inverse kinematics adopted from Chen et al.~\cite{chen2023gsdf}. We can observe removing the intermediate joint regression only drops very little on the performances. Removing the MANO regression drops slightly more in PAMJE since there is no constraint for the hand shape in the intermediate joints regression. To improve that, we add the MANO shape regression in the last variant and use the inverse kinematics to compute MANO pose parameters from the intermediate joints, which can are passed into MANO network to regress the hand mesh. We can see the performance is almost comparable with our current regressor.

\textbf{Comparable performance with some variants.}
Here, we want to emphasize that the design logic is the most important contribution of each component in our field learning module. The comparable variants share the same key ideas with our module design. For example, {\it Field gradient} also samples points near the surface (Table \ref{tab:sampling_way}), while {\it w density concatenation} also introduces distance-to-density \cite{or2022stylesdf} for SDF information encoding (Table \ref{tab:fea_aug}). They were (our) intermediate designs to the final proposed module and lacked either efficiency or performance.

\textbf{Robustness with different numbers of sampled points.} As mentioned in Sec.~\ref{sec:details}, we sample $N_v^2/n_h=600$ hand query points and $N_v^2/n_o=200$ object query points with a discretization size of $N_v=64$. Here, we sample different numbers of query points with different discretization sizes to verify that HOISDF is robust to a wide range of point sampling numbers (Fig.~\ref{fig:sampled_points}). We found that HOISDF is robust for reasonable numbers of query points. When increasing the number of query points for a discretization size of 48 one will sample many points that are far away from the hand/object, which results in large errors.

\section{Inference Speed}
\label{sec:speed}

Benefiting from the efficient way of using the field information in our field-guided pose regression module, our model can achieve real-time inference speed (30.7 FPS) on a single NVIDIA TITAN RTX GPU, which includes 10.6ms for image feature extraction, 11.5ms for query points sampling, and 10.9ms for pose attention and regression.

\section{Additional results}

\begin{figure}[ht!]
\begin{center}
\includegraphics[width=0.45\textwidth]{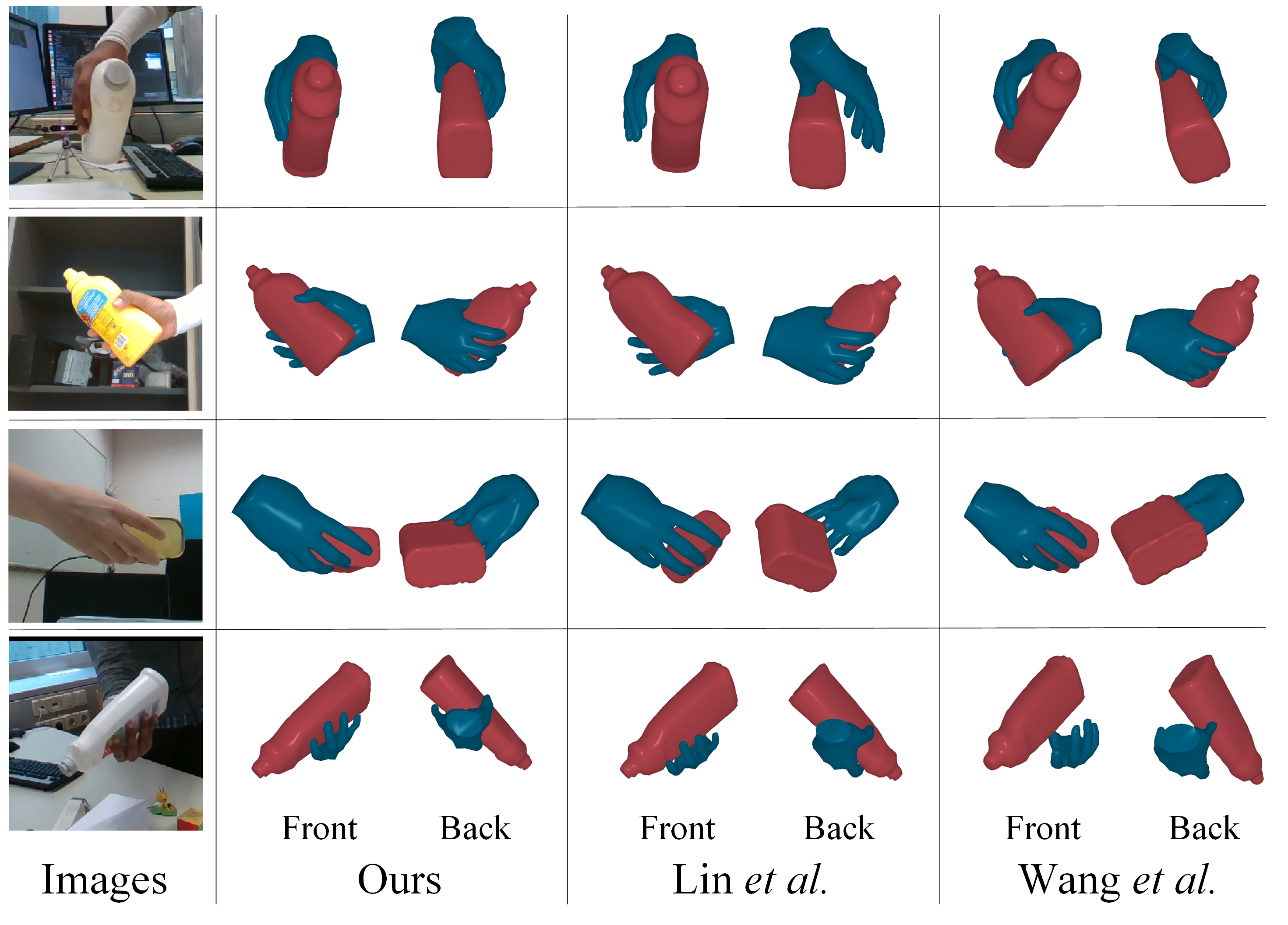}
\end{center} 
\vspace{-.5cm}
\caption{{\bf Qualitative comparisons} on the HO3Dv2 test set with Lin et al.~\cite{lin2023harmonious} and Wang et al.~\cite{ wang2023interacting}. HOISDF can produce better hand-object poses under various hand object interactions.}
\label{fig:qualitative_ho3d}
\end{figure}

\subsection{Qualitative comparison on HO3Dv2 dataset}

We visualize qualitative comparison with SOTA methods (\cite{lin2023harmonious, wang2023interacting}) on the DexYCB dataset in Sec.~\ref{sec:sota_comparison}. To further verify the effectiveness of HOISDF, we also show the qualitative comparison with the SOTA methods (\cite{lin2023harmonious, wang2023interacting}) on the HO3Dv2 dataset (Figure~\ref{fig:qualitative_ho3d}). We can observe consistent improvements in HOISDF over the SOTA methods.

\subsection{Per-object performances}

We compare HOISDF with Wang et al.\cite{wang2023interacting} that has SOTA object performances for every object category on DexYCB test set (Table~\ref{tab:perobj_dexycb}) and HO3Dv2 test set (Table~\ref{tab:perobj_ho3d}). We can observe that HOISDF outperforms Wang et al.~\cite{wang2023interacting} on almost all the object categories and all the metrics, which proves the effectiveness of our model for various objects.

\begin{table}[tp]
	\scriptsize
	\begin{center}
		\centering
    \setlength{\tabcolsep}{2.7mm}
    \begin{tabular}{cccccc}
			\toprule
			\textbf{Methods} & \multicolumn{2}{c}{HOISDF (ours)} && \multicolumn{2}{c}{Wang \emph{et al.} \cite{wang2023interacting}} \\
			\cline{2-3}\cline{5-6}
			\textbf{Metrics in [mm]} &  OME  & ADD-S  &&  OME & ADD-S  \\
			\midrule
            006\_mustard\_bottle  & 42.6 & \textbf{11.8} && \textbf{36.5} & 16.3 \\
            010\_potted\_meat\_can  & \textbf{39.7} & \textbf{14.5} && 48.6 & 22.1 \\
            021\_bleach\_cleanser  & \textbf{29.5} & \textbf{15.1} && 44.7 & 20.7 \\
			\midrule 
            Mean & \textbf{35.5} & \textbf{14.4} && 45.5 & 20.8 \\
            \bottomrule
		\end{tabular}
	\end{center}
    \vspace{-0.5cm}
	\caption{\textbf{Per-object performance on HO3Dv2 testset.} HOISDF can outperform Wang \emph{et al.} \cite{wang2023interacting} on HO3Dv2 dataset as well. }
	\label{tab:perobj_ho3d}
\end{table}

\begin{figure}[ht!]
\begin{center}
\includegraphics[width=0.45\textwidth]{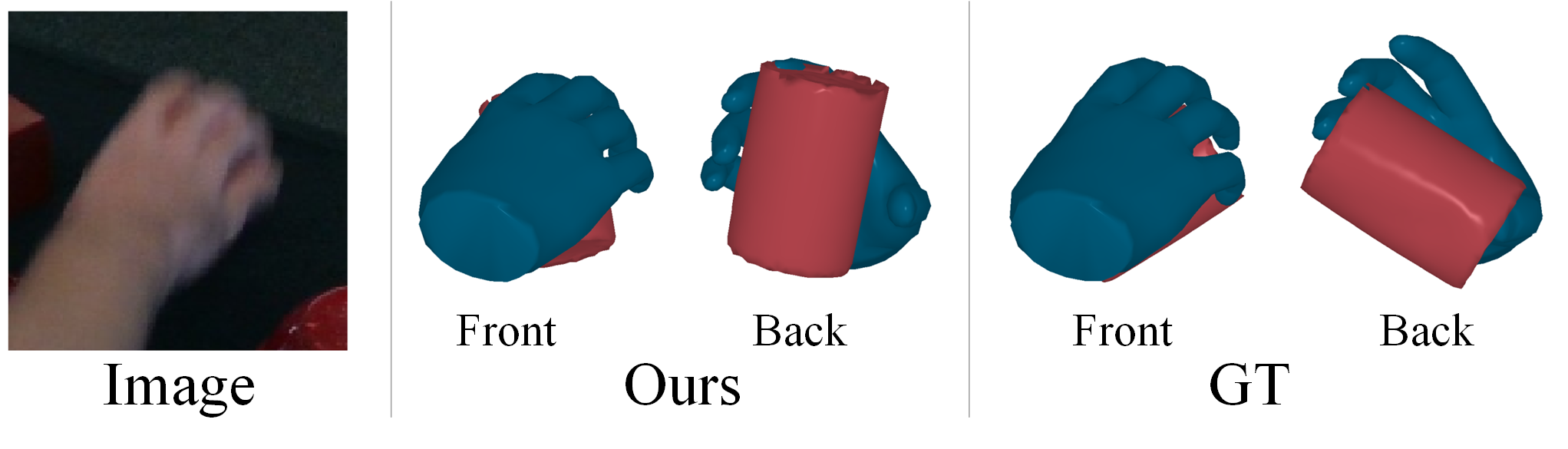}
\end{center} 
\vspace{-.6cm}
\caption{{\bf Failure case of HOISDF.} Physical plausibility could be improved. For severely occluded scenarios, the predicted hand and object meshes might intersect with each other.}
\label{fig:failure}
\vspace{-0.3cm}
\end{figure}

\subsection{Failure cases and limitations}

Although HOISDF obtains the SOTA results, it still has limitations. For severely occluded scenarios, the predicted hand and object meshes might intersect with each other (Figure~\ref{fig:failure}). Therefore, some physical constraints could be modeled during hand object pose estimation to further improve the performance.

{
    \small
    \bibliographystyle{ieeenat_fullname}

}

\end{document}